\documentclass[letterpaper, 10 pt, conference]{ieeeconf}  

\IEEEoverridecommandlockouts                              
                                                          
\let\labelindent\relax

\usepackage{graphics} 
\usepackage{overpic}
\usepackage{enumitem} 
\usepackage{overpic} 
\usepackage{color}
\usepackage{here}
\usepackage{url}
\usepackage{caption}

\title{\LARGE \bf
Deep Gesture Generation for Social Robots \\ Using Type-Specific Libraries
}

\author{
Hitoshi Teshima$^{1}$  
Naoki Wake$^{3}$  
Diego Thomas$^{1}$  
Yuta Nakashima$^{2}$ 
Hiroshi Kawasaki$^{1}$  
Katsushi Ikeuchi$^{3}$\\
$^{1}$Kyushu University \quad $^{2}$Osaka University \quad $^{3}$Microsoft \\
{\tt\small teshima.hitoshi.058@s.kyushu-u.ac.jp}
\vspace{-1mm}
}

\begin{document}



\twocolumn[{%
\renewcommand\twocolumn[1][]{#1}%
\maketitle
\begin{center}
    \centering
    \captionsetup{type=figure}
    \includegraphics[width=\linewidth]{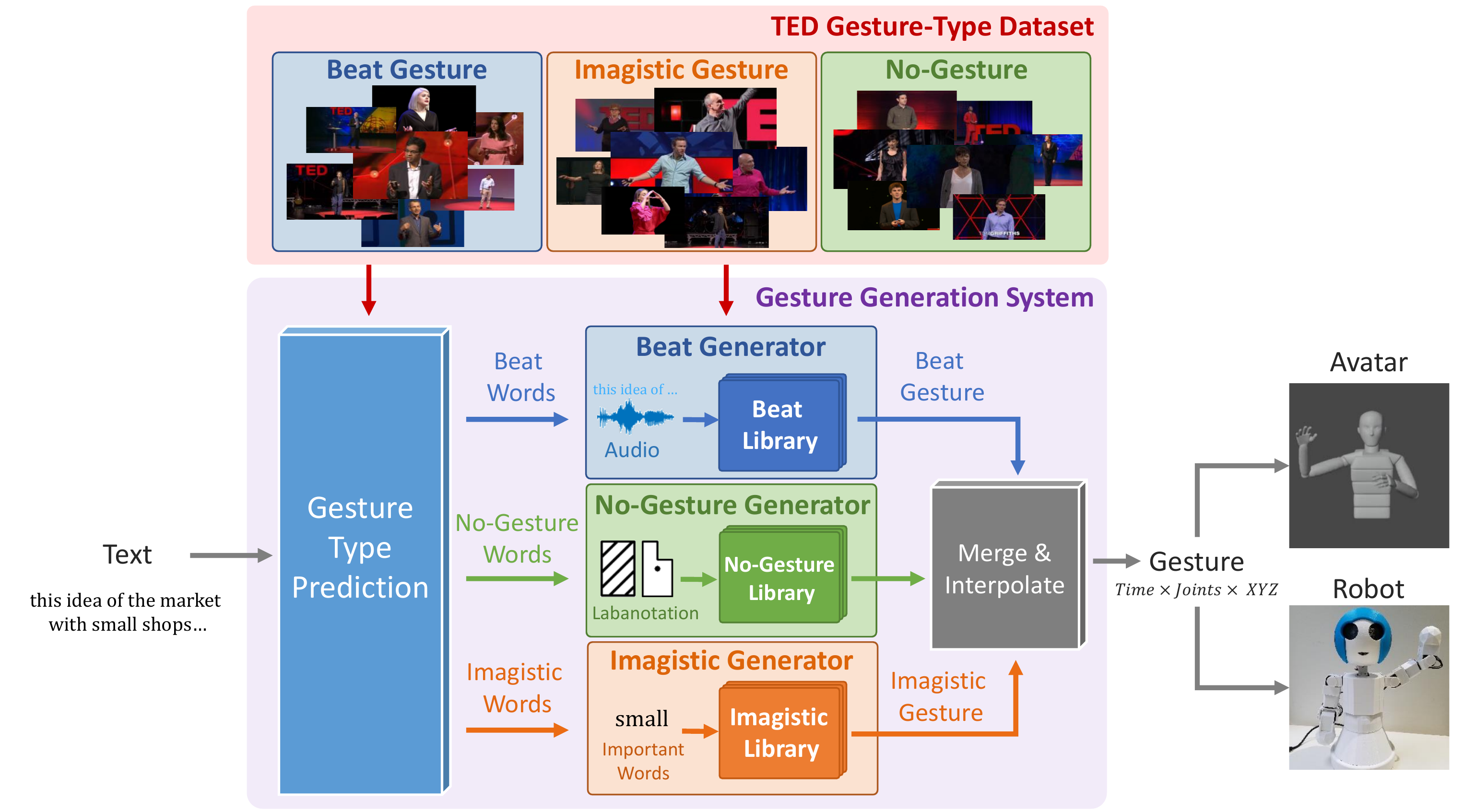}
    \captionof{figure}{Our proposed system predicts the gesture type for each word from the input text, and inputs each word sequence to separate generator to generate gestures.}
    \label{fig:overview}
\end{center}%
}]


\begin{abstract}
Body language such as conversational gesture is a powerful way to ease communication. Conversational gestures do not only make a speech more lively but also contain semantic meaning that helps to stress important information in the discussion. In the field of robotics, giving conversational agents (humanoid robots or virtual avatars) the ability to properly use gestures is critical, yet remain a task of extraordinary difficulty. This is because given only a text as input, there are many possibilities and ambiguities to generate an appropriate gesture. Different to previous works we propose a new method that explicitly takes into account the gesture types to reduce these ambiguities and generate human-like conversational gestures. Key to our proposed system is a new gesture database built on the TED dataset that allows us to map a word to one of three types of gestures: "Imagistic" gestures, which express the content of the speech, "Beat" gestures, which emphasize words, and "No gestures." We propose a system that first maps the words in the input text to their corresponding gesture type, generate type-specific gestures and combine the generated gestures into one final smooth gesture. In our comparative experiments, the effectiveness of the proposed method was confirmed in user studies for both avatar and humanoid robot.


\end{abstract}

\section{Introduction}

\label{sec:intro}

We communicate information in two ways, speech and gesture, and they are both important elements. As McNeill \cite{McNeill} argued, many human gestures are related to speech. Gestures supplement the content of the speech and make it easier for the listener to understand the information. In recent years, agents for interacting with humans, such as avatars and humanoid robots, are commonly used. If the agent communicates not only through speech but also through gestures, it will be easier for humans to understand the conversation with the agent.

Recent gesture generation systems generate gestures by learning end-to-end from text, speech, or both. However, it is still a difficult task to generate appropriate gestures that correspond to the contents of speech due to the high ambiguity of the gesture. In the workshop \cite{genea_challenge} to establish a benchmark for gesture generation, many gestures generated by the participant's method were below the Ground Truth gestures in the measure of "human-like" and below the random gestures in the measure of "gesture appropriateness". We identify the cause is that the features of text or audio as input did not match the characteristics of the gesture.

In this paper, we propose a gesture generation system with separate generators for each gesture type. The gesture types here refer to the types classified by McNeill \cite{McNeill}, which can be roughly divided into "Beat", a rhythmic gesture that appears for emphasis, and "Imagistic", a gesture that expresses something. Since Beat is a gesture like swinging arms in time with the speech, they require speech information to be generated. On the other hand, Imagistic is a gesture that express the content of the speech, such as making a circle with hands when the speaker say, "It looks like a donut". Therefore it requires semantic information to generate. In our system, at first, it predicts from the input text whether each word is likely to be Imagistic, Beat, or No-Gesture. Each predicted word sequence is then input into a generator dedicated to each gesture type. The Imagistic generator generates gestures from important words selected by DNN, and the Beat generator generates gestures from audio converted from text, using the gesture library respectively. And The No-Gesture generator generates movements that complement the movements of the previous and next gestures. The gesture libraries used in each generator are created based on the collected TED Gesture-Type Dataset. Finally, the generated gestures are interpolated and integrated to match the speech duration of the uttered word. Although the previous method \cite{teshima} also generated separate gesture types, there was not enough data to prepare the Imagisitic word extraction network, and Beat gestures were simply a small amount of pre-defined gestures. The contributions of this work are the following:
\begin{itemize}
    \item We created a dataset of TED Talks videos annotated with gesture types using crowdsourcing.
    
    \item We propose a method to generate gestures based on explicitly separating the gesture types. 
    
    \item We propose a method to build gesture libraries based on the collected data and generate each type of gestures by using the libraries.

\end{itemize}


\section{Related Work}
\label{sec:related}

\paragraph{Rule-based gesture generation}
Cassel \textit{et al} proposed a rule-based method for assigning gestures to words in conversation, according to McNeill's definition of gesture types \cite{Cassell_1}. He also created a toolkit called BEAT \cite{Cassell_2} that extracts specific words and relationships between words and assigns gestures to avatars. This method enable us to generate Imagistic gestures that match the meaning of some words. There are other rule-based systems \cite{Chien-Ming, Neff} that take text as input as well, but while these methods can generate a limited number of Imagistic gestures, they do not use speech and cannot generate Beat gestures.

\paragraph{Data-driven gesture generation}
With the remarkable development of deep learning, recent research on gesture generation tends to be data-driven methods, with some methods generating gestures from only text \cite{Yoon_1}, some from only speech \cite{Kucherenko_1, Ginosar, Ferstl}, and some from both \cite{Kucherenko_2, Yoon_2}. In recent years, the workshop \cite{genea_challenge} has been held to establish a benchmark for gesture generation. In this workshop, the evaluations of the generated gestures are significantly below the original gesture in the index of human-likeness.

Yoon \textit{et al} \cite{Yoon_1} proposed a data-driven gesture generation method that uses Seq2Seq network to translate text into gestures. However, in order to generate a gesture such as Beat, which is a gesture in which the arm and voice are synchronized, audio information is required. 

In the studies on generating gestures from audio, Kucherenko \textit{et al} \cite{Kucherenko_1} proposed a data-driven approach that uses LSTM to transform voice into gestures. Ginosar \textit{et al} \cite{Ginosar} proposed the method to generate gestures from the spectrum of speech using CNN. Ferstl \textit{et al} \cite{Ferstl} proposed a method for predicting appropriate gesture parameters from speech and using those parameters to pull gestures from a database. Using actual gestures from the database is effective in generating more realistic gestures. However, these methods using only speech as input can generate Beat, but it cannot generate Imagistic gestures to express the content of a sentence because it does not consider the context. Therefore, we use text as input and synthetic audio converted from text to generate Beat gestures and synchronize voice and hand movements. 

As studies for generating gestures using both text and speech, Kucherenko \textit{et al} \cite{Kucherenko_2} proposed a method to generate gestures by aligning the features of each and using an autoregressive network. Yoon \textit{et al} \cite{Yoon_2} proposed a gesture generation method by using a network consisting of GRU with the speaker's ID as input in addition to text and audio. These methods generate gestures directly from the end-to-end trained network, therefore they are less human-like than the actual gestures. In our method, generated gestures are expected to generate more human-like movements, since the gestures are synthesized using actual gestures in the library.

\section{TED Gesture-Type Dataset}
\label{dataset}
We introduce a new 3D gesture dataset with annotated gesture types from Yoon's TED dataset \textit{et al}\cite{Yoon_1}. The reason for using TED videos is the availability of gestures, speeches and manually annotated subtitle. The actual annotations were done on Amazon Mechanical Turk, an on-line crowdsourcing service. The annotators divided the TED video into gestures and determined whether the gesture was Beat, Imagistic, or No-Gesture. For Imagistic gestures, annotations about representative words for each gesture were also collected. 13,714 gesture sequences have been collected in total.

\begin{figure}[htbp]
    \centering
    \includegraphics[width=\linewidth]{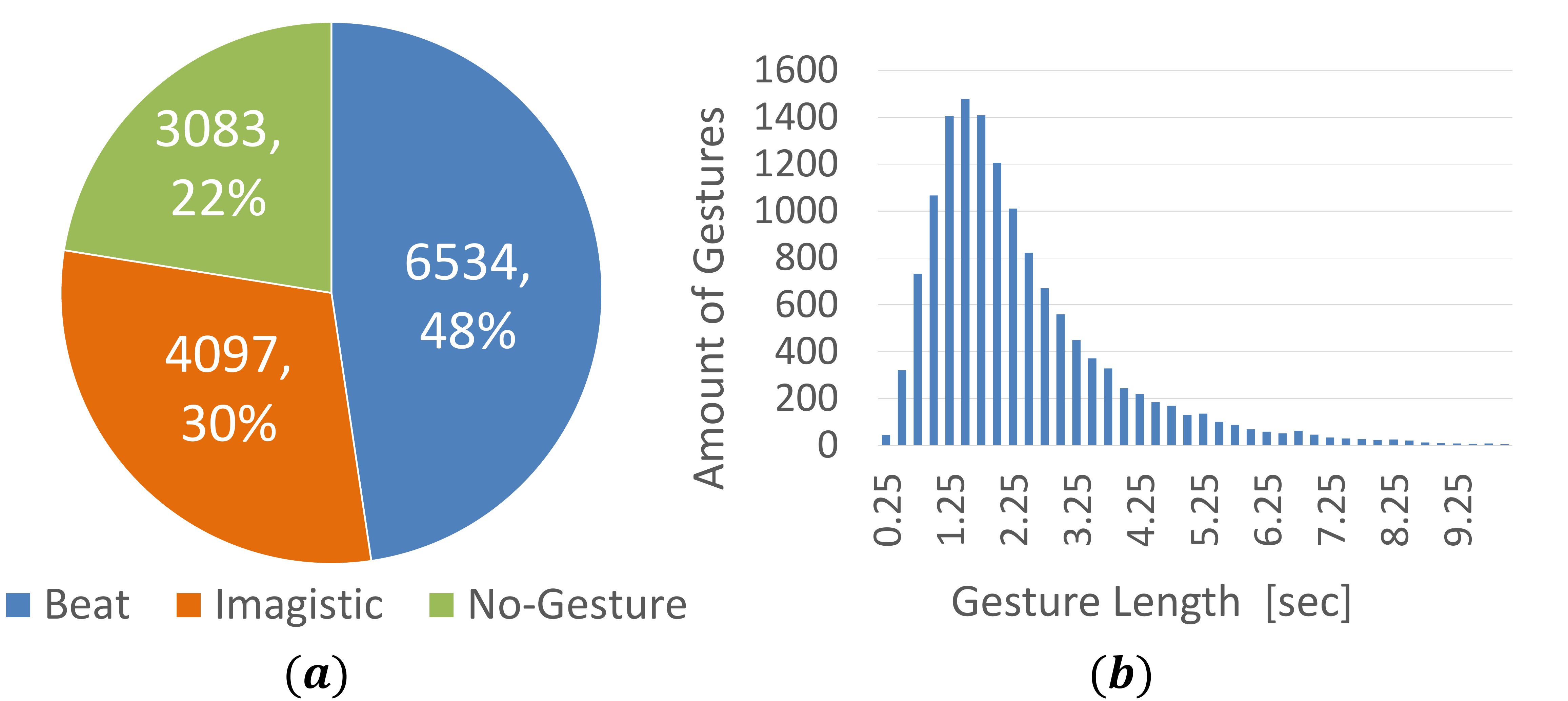}
    \caption{Contents of the TED Gesture-Type Dataset. \mbox{\boldmath $a$} Percentage of each gesture type and \mbox{\boldmath $b$} distribution of gesture length.}
    \label{fig:gesture_type_dataset}
\end{figure}

As shown in Figure \ref{fig:gesture_type_dataset}. a), Beat gestures, Imagistic gestures and No-Gestures account for 48\%, 30\% and 22\% respectively.  In the Extranarative (non-narrative speech, such as setting descriptions and character introductions) gestures of McNeil's experiment, they accounted for 54\% of Beat gestures, 28\% of Imagistic gestures, and 18\% of No-Gesture gestures, which means that the data was collected almost as defined by McNeill. Figure \ref{fig:gesture_type_dataset}. b) shows the distribution of the length of the collected gestures. Many of the gestures were 1 to 2 seconds long, and some of the Beat gestures were more than 6 seconds long.

Following the procedure of Yoon \textit{et al.}\cite{Yoon_1}, for each collected data we obtained the 2D joint coordinates from the gesture videos with using OpenPose \cite{openpose}. Then, each sequence of estimated 2D joints were lifted to 3D by using the DNN-based method proposed in \cite{3dbaseline}. The finger coordinates were not estimated due to the lack of accuracy of the estimated 3D joints. Therefore, the dataset contains segmented gesture videos included the audio, 3D joint coordinates of upper body, aligned speech text, and gesture types. We will make this TED Gesture-Type Dataset publicly available upon acceptance.

\section{Gesture Generation}
Our gesture generation system first predicts the gesture types that are likely to appear for each word, and then generates a gesture for each gesture type using a gesture library. Finally, the gestures are merged by linear interpolation. In this section, we describe how to predict gesture types \S \ref{subsec:gesture_type_prediction}, and how to use that library to generate gestures \S \ref{subsec:generating_gesture}.

\subsection{Gesture Type Prediction}
\label{subsec:gesture_type_prediction}
We broadly classify the gesture types into Beat, Imagistic and No-Gesture, referring to McNeill's definition\cite{McNeill}. This is because Beat uses audio information as input and Imagistic uses text information as input to generate gestures. For predicting the gesture types from the input text, we trained a classifier using the TED Gesture-Type Dataset introduced in Section \ref{dataset}. We used the word embedding method BERT \cite{BERT} for the word representation model. This network predict whether Beat, Imagistic, or No-Gesture is likely to appear for each word, and to smooth the prediction results, we applied a sliding window every five words. The details of this gesture type prediction are shown in previous study \cite{teshima}. We trained on a dataset that extends by a factor of 6.

\subsection{Gesture Type Specific Generators}
\label{subsec:generating_gesture}
\paragraph{Imagistic Generator}
The Imagistic gesture library consists of clusters of gestures and a list of words that represent the gestures. The details of the Imagistic gesture library are shown in previous study \cite{teshima}. 

\begin{figure}[htbp]
    \centering
    \includegraphics[width=\linewidth]{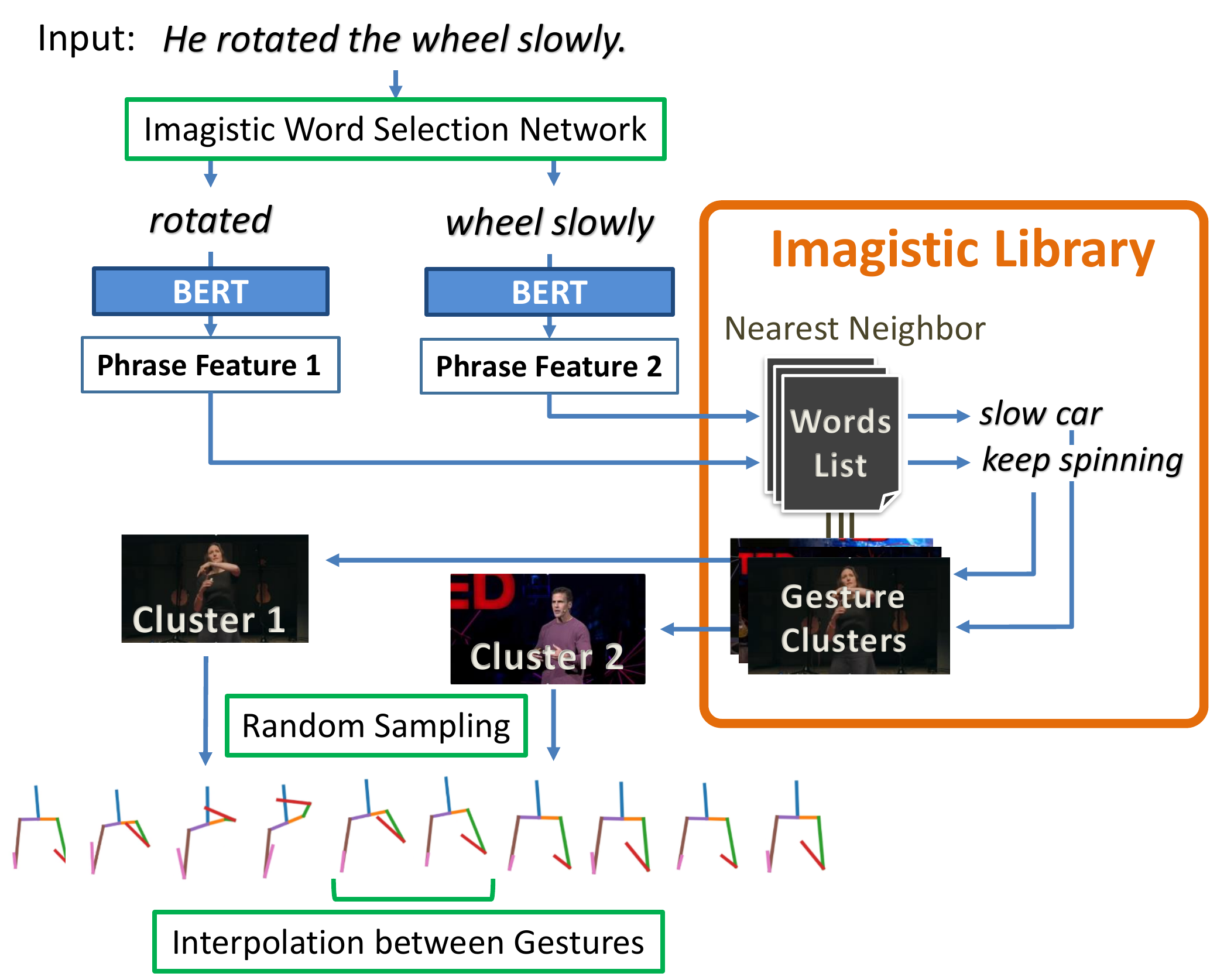}
    \caption{Flow to generate Imagistic gesture.}
    \label{fig:imagistic_generator}
\end{figure}

Imagistic gestures are generated following the flow shown in Figure \ref{fig:imagistic_generator}. First, when text is input to the module, the module select words in the text where the Imagistic gesture is likely to appear (Imagistic Word Selection Network in Figure \ref{fig:imagistic_generator}) Each word is then vectorized by the pre-trained BERT \cite{BERT} and fed into the gesture library. Next, the gesture clusters corresponding to the list of words that are nearest to the words are consulted and gestures are randomly sampled from each cluster. Finally, the sampled gestures are combined by linear interpolation. Interpolation times are so short that even simple linear interpolation looks human-like.

\begin{figure}[htbp]
    \centering
    \includegraphics[width=\linewidth]{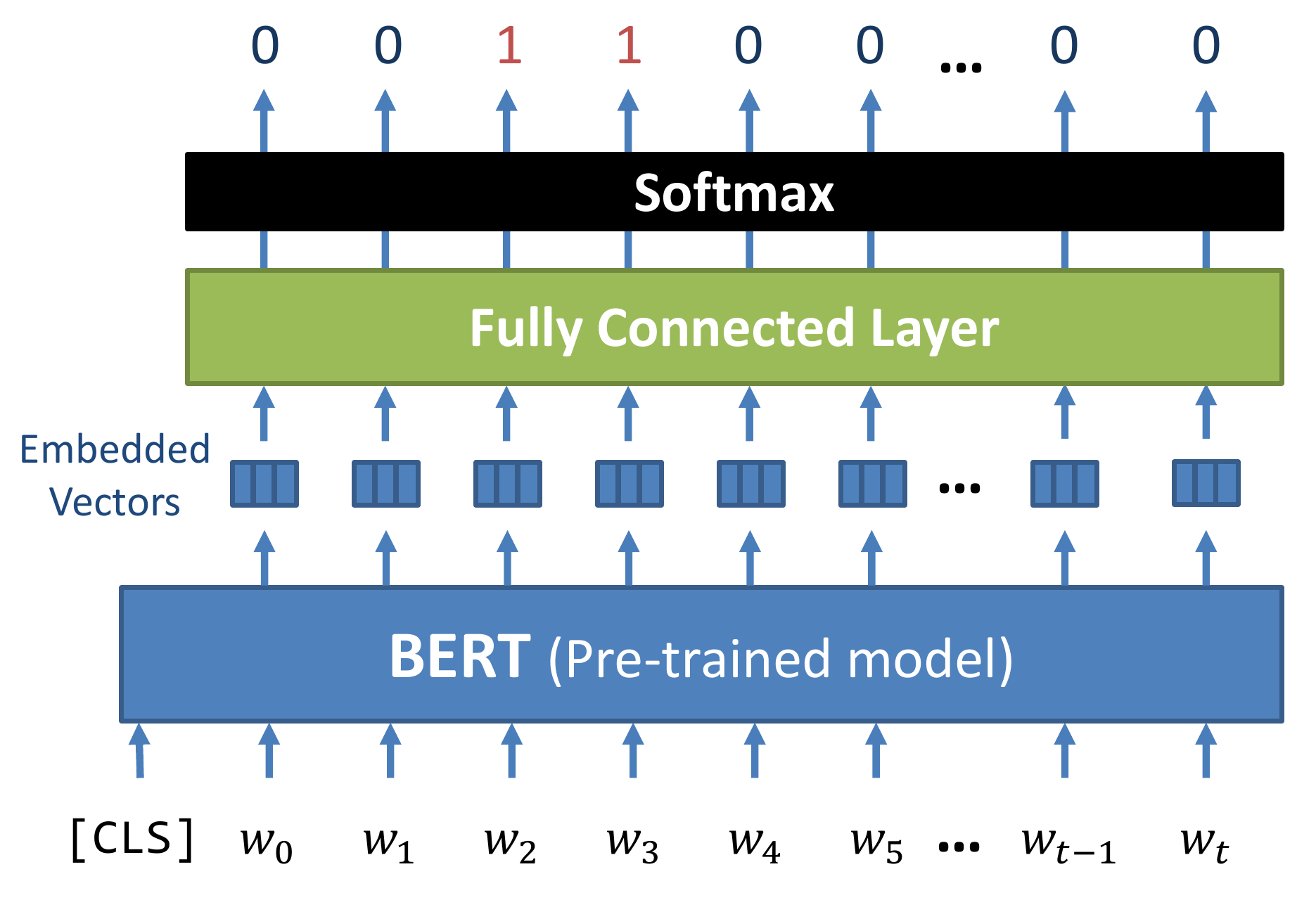}
    \caption{Imagistic Word Selection Network. This network predicts the probability that the Imagistic gesture is likely to appear in each word.}
    \label{fig:imagistic_word_selection}
\end{figure}

In this study, we use deep learning to extract the important words. The structure of the network is similar to the network used for gesture type prediction, as shown in Figure \ref{fig:imagistic_word_selection}, with the difference that the final output is a binary classification of important words or not. For the loss function, we used Binary Cross Entropy Loss and trained it to output a number between 0 and 1 for each word to indicate how important it is. For the dataset, we used the TED Gesture-Type Dataset, which also collects information on which words represent a gesture when it is annotated as an imagistic gesture. The amount of data is 1000 pieces of text data for training, 112 pieces for validation, and 279 pieces for testing.

\paragraph{Beat Generator}
As shown in Figure \ref{fig:overview}, words predicted as Beat gestures by gesture type prediction are input to the Beat Generator, which generates Beat gestures. In order to generate Beat gestures, audio information is required, not semantic information. Therefore, before generating the Beat gesture, the input text is converted into audio.

\begin{figure}[htbp]
    \centering
    \includegraphics[width=\linewidth]{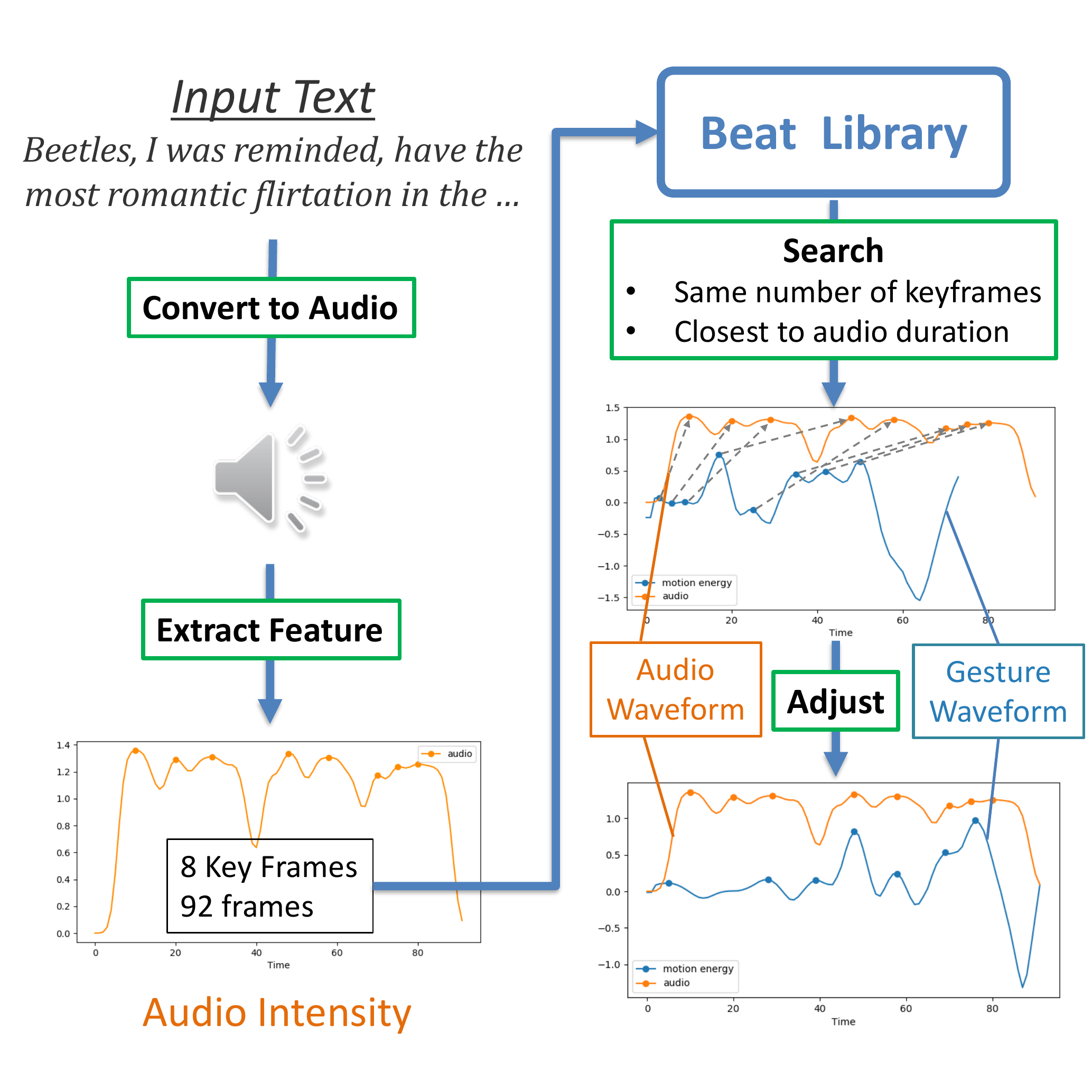}
    \caption{Flow to generate Beat gesture.}
    \label{fig:beat_generator}
\end{figure}

The flow of Beat gesture generation is shown in Figure \ref{fig:beat_generator}. The input text is first converted into synthetic speech. This process is only for text-only input; if actual human speech is available, it will be used. Intensity is extracted as a feature from the the synthetic speech, and the local maximum value of the distribution obtained by applying a Gaussian filter is used as the keyframe of the speech. The Beat gesture library consists of the 6534 Beat gestures from the TED Gesture-Type Dataset and keyframes of the gestures. The keyframes are extracted as local maxima of the Motion Energy distribution of both hands \cite{Ikeuchi2018}. The Beat gesture library is fed the number of audio keyframes and the number of audio frames, and the gestures with the same number of keyframes and the length closest to the number of audio frames in the Beat gesture library are extracted. Based on the distribution of the Motion Energy and the intensity of the audio, the gestures are slowed or quickened to match the time of each keyframe. By using these Beat gestures whose motion is converted to match the audio, we can generate gestures that are more human-like than those generated by deep learning, and that synchronize arm movements with the audio.

\paragraph{No-Gesture Generator}
The No-Gesture Generator is a module that generates No-Gesture from words that are predicted to be No-Gesture gestures by gesture type prediction. In this study, we defined the No-Gesture gesture type as a gesture in which the speaker barely moves his arms even though he is speaking. In the No-Gesture Generator, as in the other generators, we built a gesture library and generated gestures by referring to No-Gesture that matched the gestures before and after and the duration of the input text.

The No-Gesture library consists of 3083 motions annotated as No-Gesture in the TED Gesture-Type Dataset, the number of frames of the gestures, and labanotations converted from the gestures. When generating the No-Gesture, the position of the labanotation of the last keyframe of the previous gesture and the first keyframe of the next gesture are referenced, and the gestures are retrieved a gesture from the No-Gesture library that matches the positions of the labanotation of the previous and next gestures. The selected No-Gesture is slowed and quickened to match the vocalization period obtained by converting the input text into speech, and the No-Gesture is generated.

\section{Experiments}
To find out how natural the generated gestures are, we conducted perceptual experiments on Amazon Mechanical Turk. This experiment was conducted using the same gesture visualizer and the same user interface, referring to previous work \cite{genea_challenge} that determined benchmarks in gesture generation. The evaluation items were also the same as in the previous work: "How human-like does the gesture motion appear?" without audio and "How appropriate are the gestures for the speech?" with audio. After watching each gesture, the participants rated each question on a scale of 100.

The dataset used for the evaluation was the Trinity Speech Gesture Dataset \cite{trinity_dataset}, and the video of the gestures to be generated was created to be close to 10 seconds, which is the average number of seconds mentioned in the previous work \cite{genea_challenge}. In the work, the joints of the spine that can be moved are divided into five, but the gestures generated in this study conform to the joints estimated by OpenPose \cite{openpose}, and the joints of the spine do not move, so the spine of the avatar is fixed in this experiment. We used 30 different gesture videos for the evaluation.

This experiment was conducted on Amazon Mechanical Turk. The participants were selected if they satisfied the following three conditions: 1) they had completed more than 100 tasks, 2) Their approval rate of task is over 90\%, and 3) they had passed the Attention Check. Attention checks were conducted to improve the quality of the workers, similar to previous studies. 136 participants met these requirements for the question of "Human-likeness", and 124 participants met these requirements for the question of "Appropriateness".

\begin{figure*}[htbp]
    \centering
    \includegraphics[width=\linewidth]{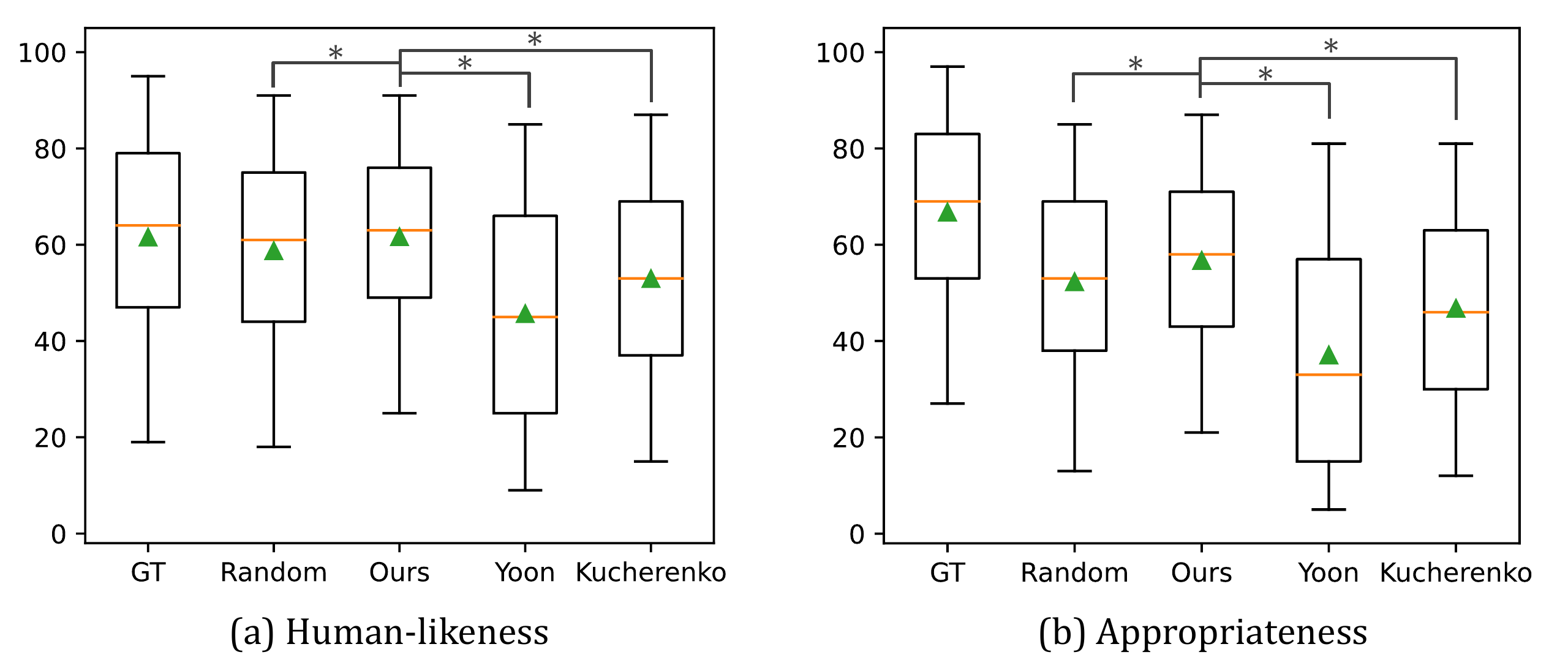}
    \caption{Results of an experiment to evaluate gestures with avatars. (*$p<0.05$) The orange line represents the median, and the green triangle represents the mean. Box edges are at 25 and 75 percentiles, while whiskers cover 95\% of all ratings for each system.}
    \label{fig:eval_genea}
\end{figure*}

The gestures we compared were GT gestures (GT) from Trinity Speech Gesture Dataset, randomly assigned gestures from the TED gesture database (Random), and two gesture generation methods from other previous studies. The two gesture generation methods compared are a method proposed by Yoon \textit{et al.} \cite{Yoon_2} (Yoon) that uses GRU to generate gestures from text, speech, and speaker ID, and Kucherenko \textit{et al.}'s method \cite{Kucherenko_2} (Kucherenko) that uses BERT and recursive networks to generate gestures from text and speech. Since Yoon \textit{et al.}'s method requires the speaker's ID as input, we used a randomly chosen ID as input. These two end-to-end methods were trained using training data from the Trinity Speech Gesture Dataset.

\begin{figure}[htbp]
    \centering
    \includegraphics[width=\linewidth]{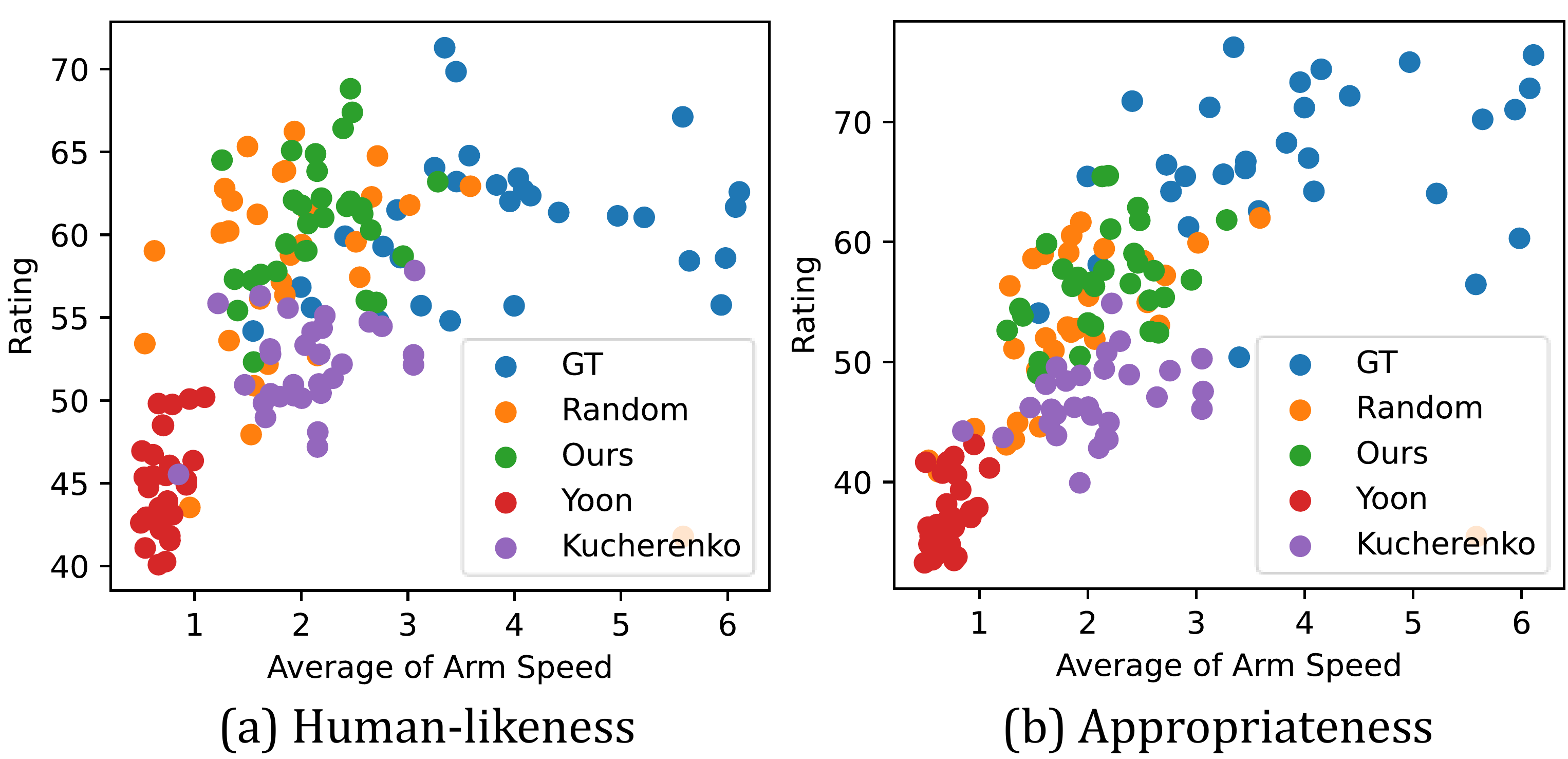}
    \caption{Relationship between the evaluated value of each gesture and the angular velocity of the arm.}
    \label{fig:eval_genea_speed}
\end{figure}

Figure \ref{fig:eval_genea} (a) and Figure \ref{fig:eval_genea} (b) show the results of the "Human-likeness" and "Appropriateness" experiments, respectively. The results of ANOVA test showed a statistically significant difference with p-value less than 0.05. The results of Tukey's multiple comparison test showed that GT's gesture is the best rated for both measures, and our gesture is the second best with $p<0.05$. Figure \ref{fig:eval_genea_speed} shows the distribution of the evaluation value and the angular velocity of the arm for each gesture. The vertical axis shows the evaluation value in 100 steps, and the horizontal axis shows the angular velocity of the arm for the gesture. From this figure, we can conclude that Yoon et al.'s method and Kucherenko et al.'s method generated excessively smooth gestures with little movement, and thus were rated low.

\subsection{Experiments with Robot}
We evaluated the gestures generated by our system on a humanoid robot as well. The robot used in the evaluation was one that could be easily created with a DIY kit developed in a previous study \cite{teshima,gestureBotDesignKit}. This robot originally has a function to extract the nearest word in the input text using word2vec \cite{word2vec}  and output a gesture from about 90 predefined gestures. We use this gesture as a baseline and compare it to the gestures by proposal method. Since the baseline method generates only one gesture for each input text, we divided the input text into 15 word segments to generate gestures, considering the fairness due to the length of the text. Also, when we generated the robot's gestures using the proposed method, we made sure that the robot's hand always came to rest position at the end of the gesture to make it look more human-like. The input texts were randomly selected 10 from the TED dataset.

In addition to the same metrics of "Human-likeness" and "Appropriateness" as in the avatar experiment, we also evaluated the diversity of gestures. We recruited 30 participants at Amazon Mechanical Turk. They compared the baseline gestures with the gestures by proposed method and answered which is better in each metric.

\begin{figure}[htbp]
    \centering
    \includegraphics[width=\linewidth]{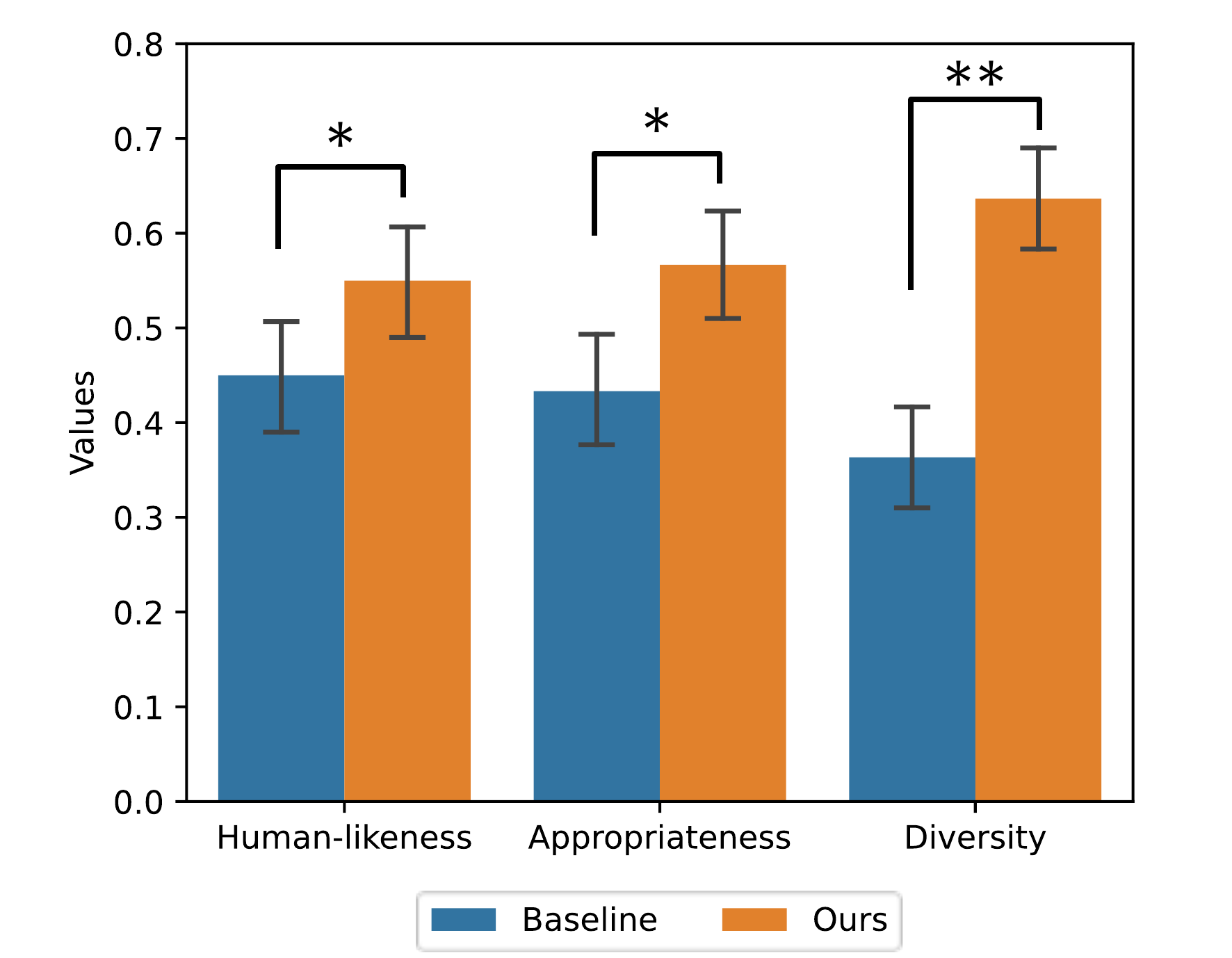}
    \caption{Results of an experiment to evaluate gestures with humanoid robot. (*$p<0.05$; **$p<0.005$)}
    \label{fig:eval_msrabot}
\end{figure}

Figure \ref{fig:eval_msrabot} summarizes the results of the evaluation with humanoid robot. Across all metrics, gestures by proposal method outperformed the baseline. Statistical tests showed that there was a significant difference in the "Human-likeness" and "Appropriateness" at p<0.05, and a significant difference in the "Diversity" at p<0.005.In particular, in terms of "Diversity", the baseline generates only a limited number of gestures, about 90, which show the richness of our gesture library.

\section{Conclusions}
We proposed a method for generating gestures from speech text that explicitly separates each gesture type. We created a dataset of TED Talks videos annotated with the types of gestures, and used it to predict gesture types from text and to build a gesture library. When generating Beat, the input text was converted to audio before generating it, and when generating Imagistic, the gestures were generated by extracting important words from the input text. Each generator can generate more human-like gestures than those generated by end-to-end DNNs by using gesture libraries built from the TED Gesture-Type Dataset. In experiments with avatars, our gestures outperformed end-to-end DNN-generated gestures and randomly selected gestures in both the "Human-likeness" and "Appropriateness" metrics. In our experiments with the robot, our gestures outperformed the baseline gestures, showing the diversity of our gesture library.

As a limitation, the accuracy of the gesture type prediction and the extraction of important words in the Imagistic generator is still insufficient, and there is room for improvement. Also, since the gesture data conforms to the OpenPose \cite{openpose} joint definitions, the spine and finger joints do not move in our generated gestures. Therefore, for future work, we can expect to express gestures even with finger joints and generate facial expressions to make communication with avatars and robots more understandable.

\end{document}